\documentclass{article}

\usepackage{graphicx}
\usepackage{comment}
\usepackage{amsmath,amssymb} % define this before the line numbering.
\usepackage{color}
\usepackage{bbold}
\usepackage{subcaption}
\usepackage{caption}
\usepackage{xcolor}
\usepackage{subcaption}
\usepackage{multirow}
\usepackage{wrapfig}
\usepackage{caption}
\usepackage{subcaption}
\usepackage[final]{corl_2020} % Uncomment for the camera-ready ``final'' version.

\DeclareMathOperator*{\argmax}{arg\,max}

\title{Learning Object-conditioned Exploration using
Distributed Soft Actor Critic}

\author{
  \textbf{Ayzaan Wahid, Austin Stone, Kevin Chen, Brian Ichter, Alexander Toshev}\\
  Robotics at Google
}

\begin{document}
\maketitle

%===============================================================================

\begin{abstract}
Object navigation is defined as navigating to an object of a given label in a complex, unexplored environment. In its general form, this problem poses several challenges for Robotics: semantic exploration of unknown environments in search of an object and low-level control. In this work we study object-guided exploration and low-level control, and present an end-to-end trained navigation policy achieving a success rate of 0.68 and SPL of 0.58 on unseen, visually complex scans of real homes. We propose a highly scalable implementation of an off-policy Reinforcement Learning algorithm, distributed Soft Actor Critic, which allows the system to utilize 98M experience steps in 24 hours on 8 GPUs. Our system learns to control a differential drive mobile base in simulation from a stack of high dimensional observations commonly used on robotic platforms. The learned policy is capable of object-guided exploratory behaviors and low-level control learned from pure experiences in realistic environments. 
\end{abstract}


\section{Introduction}\label{sec:intro}

Visual Navigation is a long-standing problem at the intersection of Robotics and Computer Vision. Much of the existing effort in the field has been focused on safely and efficiently navigating a robot to a geometrically specified goal, commonly known as point goal navigation. However, in many practical settings the goal location is unknown, such as in solving the problem: ``Where are my keys?'' This class of problems, in which the goal is specified semantically, is known as Semantic Visual Navigation. Tackling this problem involves overcoming numerous challenges, such as generating intelligent plans to efficiently explore the space and controlling the robot in a safe and efficient manner. The vision community has studied visual navigation using high-level discrete control with a known goal in an unknown environment, whereas the robotics community has investigated using low-level continuous control to reach a \textit{known} goal in a \textit{known} environment. We seek to combine these, and learn to navigate to a target of unknown location with low-level continuous control in an unknown environment.
% What is ObjectNav and why?
% Robot Navigation is a long standing problem at the intersection of Robotics and Computer Vision. It touches on both understanding of the space around us using visual inputs as well as precise control of a kinematic systems under the laws of physics. As there are many aspects to this problem, there isn't one accepted definition of navigation. A common version is point goal navigation where the goal is specified geometrically. Arguably, metric navigation requires limited understanding of surrounding semantics and exploration behaviors. Furthermore, it has been studied in the computer vision community~\cite{savva2019habitat,chaplot2020learning} with learned policies primarily in the unrealistic setting of discrete control. In the Robotics community there is a larger body of work on this problem where a focus is low-level continuous control. A different version of navigation is Semantic Visual Navigation, where the target is specified semantically. In this setup, the navigation policy has to perform exploration, as the target object location is generally unknown and might be outside the field of view.

\begin{figure}[t]
        \centering
        \includegraphics[width=0.7\textwidth]
        {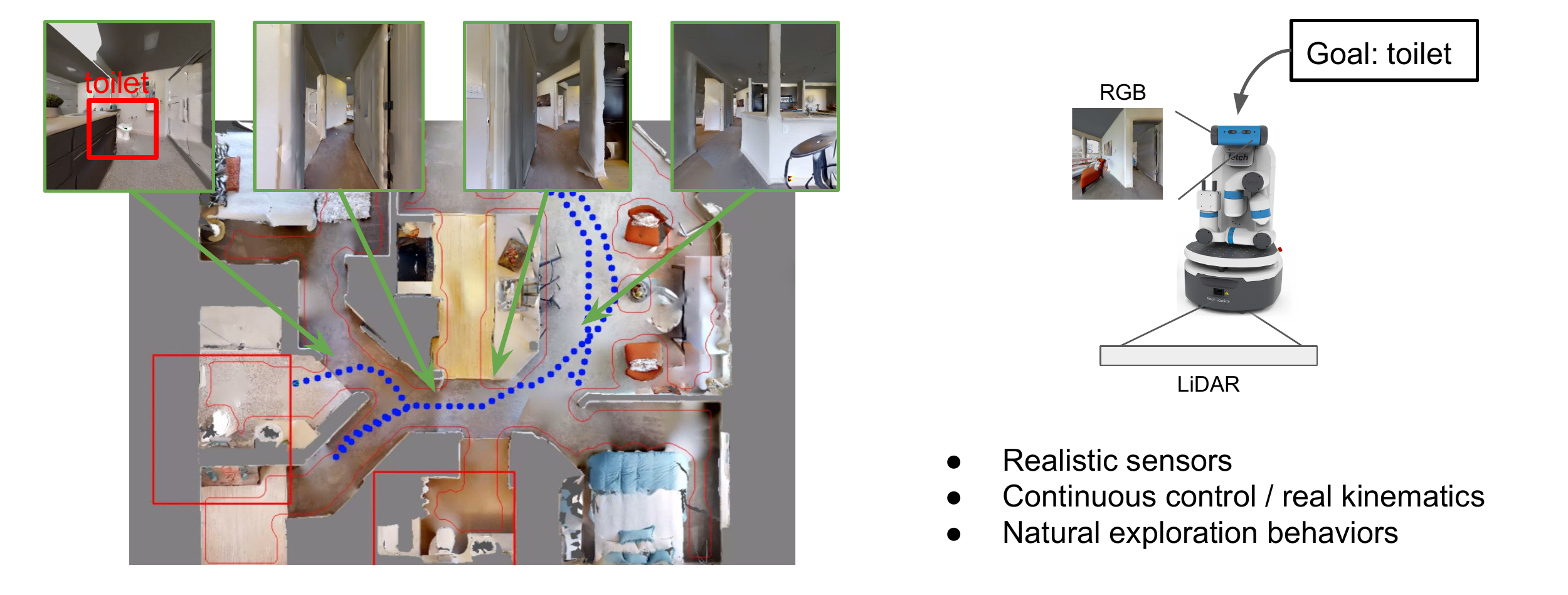} \caption{\label{fig:intro_fig}Navigating to an object requires object-conditioned exploration (e.g.,~efficiently exploring a space using contextual cues) and low level control (e.g.,~controlling a differential drive). We present a scalable distributed SAC system to learn an object navigation policy in realistic environments, addressing all of the above challenges.}
\end{figure}

In this work we focus on object-conditioned exploration and low-level control, for the purpose of navigating to objects in realistic simulated settings (see Fig.~\ref{fig:intro_fig}). We assume no prior knowledge of the environment and no localization information. Further, we aim towards a simulation setup which mimics many of the challenges present in real world robotics. We use environments based on scans of real homes~\cite{xia2018gibson} to provide maximum visual realism and complexity, and simulate the physics of a differential drive robot to address the challenges of low-level control. While we primarily tackle the \textit{exploration} and \textit{control} components of this problem, we show results that demonstrate the challenges of learning policies which can perform exploration, recognition, and control in search of an object. We leave real world deployment for future work, as this would require more work in sim-to-real transfer or training in the real, which are challenging due to the sim-to-real gap and limited ability to collect large amounts of data on a robot. However, the more realistic setup we provide in simulation is a step towards better real world performance compared to prior methods.

For the above setup of semantic navigation, we revisit end-to-end learned navigation policies. We introduce a distributed version of the Soft Actor Critic (SAC) algorithm~\cite{haarnoja2018soft_constrained} which allows us to train a policy from experience. SAC, as an off-policy reinforcement learning algorithm, is amenable to scalable parallelization, where we parallelize both the model optimization and the experience collection. This allows us to train a conventional Neural Net (ResNet50 + LSTM) as a policy to output continuous actions from rich, high dimensional, realistic robot observations. We also note that we specifically choose an off-policy algorithm, since these are more sample efficient than on-policy alternatives. Sample efficiency is important in the robotics domain, where data collection is expensive and maximal data re-use is highly desired. Using this system, we achieve a success rate of 0.68 and SPL of 0.58 for navigating to $9$ different object types by training over 98 million experience steps per 24 hours on 8 GPUs.

\begin{figure}
\centering
\begin{minipage}{0.65\textwidth}
\begin{subfigure}{.42\textwidth}
\includegraphics[width=\textwidth]{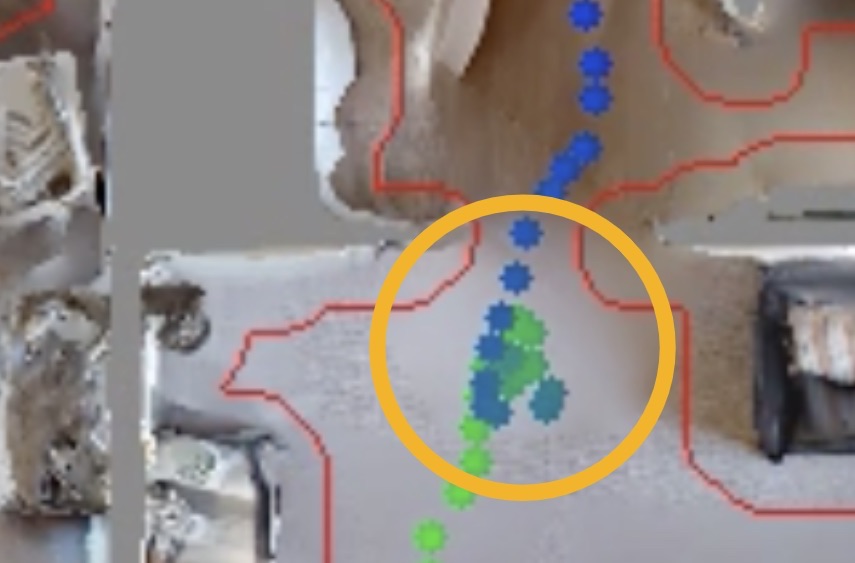}
\caption{Circle}\label{fig:behaviors_circling}
\end{subfigure}
\begin{subfigure}{.403\textwidth}
\includegraphics[width=\textwidth]{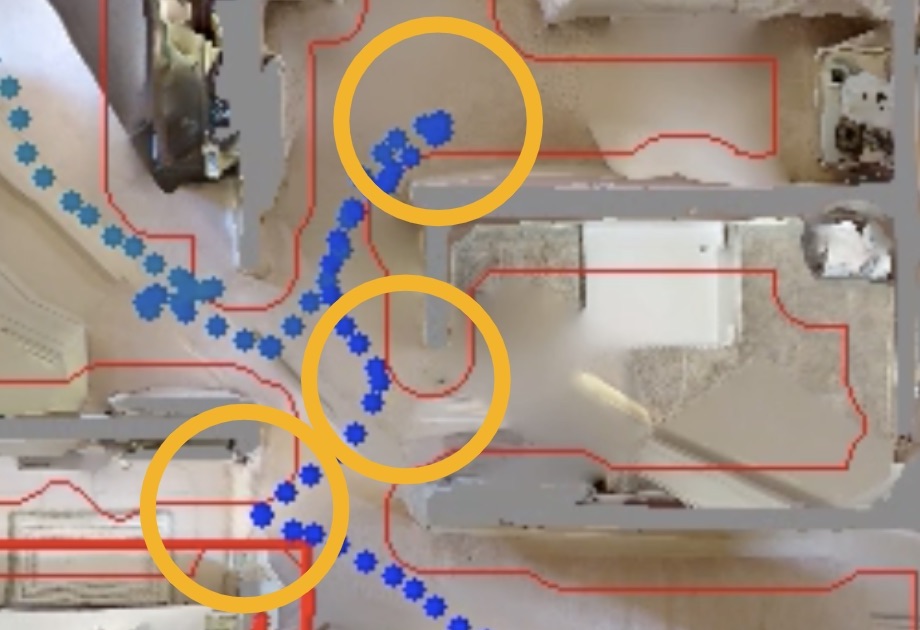}
\caption{Peek}\label{fig:behaviors_peeking}
\end{subfigure}
\begin{subfigure}{.158\textwidth}
\includegraphics[width=\textwidth]{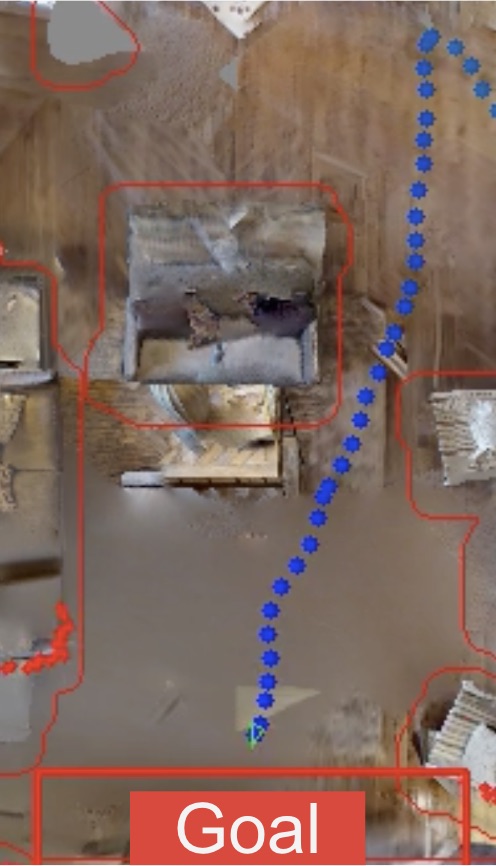}
\caption{Beeline}\label{fig:behaviors_beelining}
\end{subfigure}
\end{minipage}
\caption{
(\ref{fig:behaviors_circling}) The agent learns to move straight and periodically turn in place to gather information.
(\ref{fig:behaviors_peeking}) The agent learns to ``peek'' into rooms and backtrack if the object is unlikely to be present.
(\ref{fig:behaviors_beelining}) The agent learns to move directly to the goal once it is seen.
}
\label{fig:behaviors_intro}
\vspace{-0.6cm}
\end{figure}

We demonstrate that a single end-to-end model can learn several important skills in the realistic and visually rich simulation environments required for object navigation. First the agent learns natural exploration behaviors, such as turning in place when the object is not visible, backtracking when failing to find the goal object, and going straight to the target object when visible (Fig.~\ref{fig:behaviors_intro}). These behaviors are then conditioned on the type of object, so that the agent performs different exploration behaviors for different types of objects. It is important to note that these emerge from experience without any explicit supervision.

Finally, we demonstrate the challenges of using low-level control, and the utility of having a comprehensive set of sensors, e.g., range sensing, which allows for more collision free object navigation.

\section{Related Work}\label{sec:rel-work}
The proposed work is related to different approaches towards visual navigation. Classical techniques for navigation typically start with map construction, followed by planning and execution of the planned trajectories. The mapping stage uses visual observations to simultaneously localize and construct maps, either topological or semantic and metric (see for an extensive review ~\cite{SLAM16,armeni_cvpr16} of SLAM).

\textbf{Learned Navigation System} In the last few years a large body of work on learned navigation policies has emerged~\cite{Learning2Navigate,gupta2017cognitive,zhu2017target,mishkin2019benchmarking}. This body of work has laid the groundwork for learning-based navigation, where a sensors-to-control policy is learned from experience or demonstrations. Researchers have studied a wide variety of navigation problems, such as metric navigation~\cite{anderson2018evaluation,sax2019learning,wijmans2019dd,chaplot2020learning}, target-driven semantic navigation~\cite{zhu2017target,yang2018visual,mousavian2019visual}, and exploration~\cite{zhang2017neural,chaplot2020learning,chen2018learning,parisotto2017neural,savinov2018semi,fang2019scene}. 
%, and visual question answering~\cite{IQAGordon,das2018embodied,wijmans2019embodied}.

%Most relevant to our problem setup, Yang et al.~\cite{yang2018visual} train Graph Convolutional Networks with A3C~\cite{A3CAtari16} for object navigation, working only with discrete commands. Mousavian et al.~\cite{mousavian2019visual} address a similar problem using DaGGER~\cite{ross2011reduction}, a form of learning from demonstrations.

Most relevant to our problem setup, Yang et al.~\cite{yang2018visual} train Graph Convolutional Networks for object navigation with discrete commands. Mousavian et al.~\cite{mousavian2019visual} address a similar problem using learning from demonstrations.

%Although the above line of work presents a strong and promising direction, it oftentimes falls short of being realistic. In particular, researchers make various assumptions which simplify the setup such as the existence of oracle semantic segmentation, oracle localization, and discrete actions. As a result, such systems do not address all of the challenges present in the real world. Further, many of these approaches are trained and evaluated on environments of limited realism, such as SUNCG \cite{song2016ssc}, House3D~\cite{House3D_NIPS17}, AI2THOR~\cite{ai2thor}, or Doom~\cite{kempka2016vizdoom}. %, or Matterport 3D~\cite{MINOSFall17}.

More recently, we have seen two different developments which facilitate learning realistic navigation policies. First, reinforcement learning (RL) algorithms have become more stable and mature, e.g.,\cite{schulman2017proximal, haarnoja2018soft}. Second, we have seen the development of navigation simulation environments, which simulate realistic visuals and physics at scale, e.g.~\cite{Gibson_CVPR18,savva2019habitat}. In this paper, we  leverage these new simulations and learn a navigation policy which tackles both complex visuals and physics.

These advances have already been utilized, to a degree, for point navigation. Chaplot et al.~\cite{savva2019habitat} demonstrate a hierarchical learned navigation system for point navigation with very strong results on the Habitat challenge~\cite{savva2019habitat}. Sax et al.~\cite{sax2019learning} apply PPO on top of mid-level image representations for strong results on the same benchmark. For the same problem, Wijmans et al.~\cite{wijmans2019dd} demonstrate even stronger results using a distributed version of PPO.

While point navigation~\cite{savva2019habitat,anderson2018evaluation} has made remarkable progress, its applicability to real-world settings remains limited. In most applications, the precise metric location of the target is unknown, which is the case in object navigation. This important difference introduces serious challenges as the agent has to efficiently plan and explore the unknown environment. Furthermore, real-world robots operate using continuous control rather than the commonly used discrete left/right/forward actions. These important details must be taken into account in the choice and design of the algorithm. In contrast to prior works, which have tackled some of these challenges individually (Table~\ref{tab:related-work-comparison}), we propose a method to handle them altogether.% While prior works have tackled some of these challenges individually (Table~\ref{tab:related-work-comparison}), we propose a method to handle them altogether.  % Many prior works do not tackle the aforementioned challenges, as shown in Table~\ref{tab:related-work-comparison}.
% Related works: structured memory approaches (SPTM, SMT, neural map, etc.)

\textbf{Motion Planning and Obstacle Avoidance} In this work we emphasize  having a realistic setup, meaning continuous control. The main challenge is collision-free movement respecting the geometry of the robot and obstacles. Classical approaches include Artificial Potential Fields~\cite{khatib1986real} or Dynamic Window Approach~\cite{fox1997dynamic}. These require an explicit model of the environment, which we do not have at test time.

Deep Learning has found traction in Motion Planning research as well. In more detail, control policies have been trained using supervised learning~\cite{chen2015deepdriving} and DDPG~\cite{faust2018prm} for point navigation. %Similar to Chiang et al.~\cite{chiang2019learning}, our policy learns a differential drive control which avoids collisions. 
However, instead of point navigation, we achieve this in the context of semantic, long-range navigation.

% Table with General.
% \begin{table}
% \begin{center}
% \begin{tabular}{l|c|c|c|c|c|c|c|c|}
% % \hline
%   & No Goal & Cont. & Visual & General & Physics & Obj.-conditioned  \\
%   & Vector & Control & Realism & Exploration & Realism & exploration \\
% \hline%\hline
% Zhu et al.~\cite{zhu2017target} &   &   &  & \checkmark &   & \checkmark \\
% % \hline
% Gupta et al.~\cite{gupta2017cognitive} &  &   & \checkmark &  &  &  \\
% % \hline
% Mousavian et al.~\cite{mousavian2019visual} & \checkmark &  & \checkmark & \checkmark &  &  \\
% % \hline
% Sax et al.~\cite{sax2019learning} & \checkmark &  & \checkmark & \checkmark & \checkmark &  \\
% % \hline
% % Fang et al.~\cite{fang2019scene} &  &  &  & X & X &  & X & X \\
% % \hline
% Yang et al.~\cite{yang2019embodied} & \checkmark &  &  & \checkmark &  & \checkmark  \\
% % \hline
% % Georgakis et al.~\cite{georgakis2019simultaneous} & X &  & X & X & & & X & X \\
% % \hline
% Chaplot et al.~\cite{chaplot2020learning} & \checkmark &  & \checkmark &  &  &    \\
% % \hline
% Wijmans et al.~\cite{wijmans2019dd} &  &  & \checkmark & \checkmark &  &    \\
% % \hline
% Ours & \checkmark & \checkmark & \checkmark & \checkmark & \checkmark & \checkmark  \\
% %  \hline
% \end{tabular}
% \end{center}
% \caption{\label{tab:related-work-comparison}Comparison with prior learning-based visual navigation approaches.} % re-caption this
% \end{table}
\section{Object-conditioned Exploration}
\vspace{-2mm}
Object Navigation (ObjectNav) is defined as the problem of navigating to an object, specified by a label, in unstructured and unknown environments~\cite{anderson2018evaluation}. In this work, we want to tackle object-conditioned exploration in its most challenging form:

\textbf{Realistic complex visuals:} the agent navigates in realistic environments. We propose to use GibsonEnv~\cite{xia2018gibson} as a set of scans of real homes.

\textbf{Realistic physics:} the agent has realistic embodiment. We suggest using a model of a robotic system -- more precisely a differential drive with mounted RGB, depth, and 1-D LiDAR modeled after a Fetch robot.
    
\textbf{Semantics:} We use $9$ object types to navigate to. The starting location can be up to $15$m away, and as such the robot explores multiple rooms before getting to the target.

The problem of object navigation brings together a range of skills -- visual recognition, object-conditioned exploration of unknown environments, and low-level control. As such, in this study we mostly assume that the goal object detection is given by an oracle. 

\begin{table}[t]
\centering
\footnotesize{
\begin{tabular}{l|c|c|c|c|c|c|c|c|}
% \hline
  & No Goal & Cont. & Visual  & Physics & Obj.-cond.  \\
  & Vector & Control & Realism  & Realism & exploration \\
\hline%\hline
Zhu et al.~\cite{zhu2017target} &  \checkmark &   &   &   &  \\
% \hline
Gupta et al.~\cite{gupta2017cognitive} &  &   & \checkmark  &  & \checkmark \\
% \hline
Mousavian et al.~\cite{mousavian2019visual} & \checkmark &  & \checkmark  &  & \checkmark \\
% \hline
Sax et al.~\cite{sax2019learning} & \checkmark &  & \checkmark & \checkmark &  \\
% \hline
% Fang et al.~\cite{fang2019scene} &  &  &  & X & X &  & X & X \\
% \hline
Yang et al.~\cite{yang2019embodied} & \checkmark &  &   &  & \checkmark  \\
% \hline
% Georgakis et al.~\cite{georgakis2019simultaneous} & X &  & X & X & & & X & X \\
% \hline
Chaplot et al.~\cite{chaplot2020learning} & \checkmark &  & \checkmark   &  &    \\
% \hline
Wijmans et al.~\cite{wijmans2019dd} &  &  & \checkmark  &  &    \\
\hline
Ours & \checkmark & \checkmark & \checkmark  & \checkmark & \checkmark  \\
%  \hline
\end{tabular}}
\caption{\label{tab:related-work-comparison}Comparison with prior learning-based visual navigation approaches.} % re-caption this
\vspace{-0.8cm}
\end{table}

\section{Learning of General Navigation Policy}
\vspace{-2mm}
Consider the case where the robot starts in the dining room and is looking for an oven. The robot may perform rotational motion to explore the environment, spot the kitchen bar counter in the distance, and use this cue to move towards it, hypothesizing that there is a kitchen area with the goal behind the counter. This type of reasoning process may benefit from end-to-end neural network policies which can tightly integrate these skills. Furthermore, these policies may benefit from learning via experience in realistic environments, since explicit supervision for the above reasoning is hard to define.

\subsection{Distributed Soft Actor-Critic}
\vspace{-2mm}

\begin{figure}[!htb]
        \center{\includegraphics[width=0.6\textwidth, height=0.2\textwidth]
        {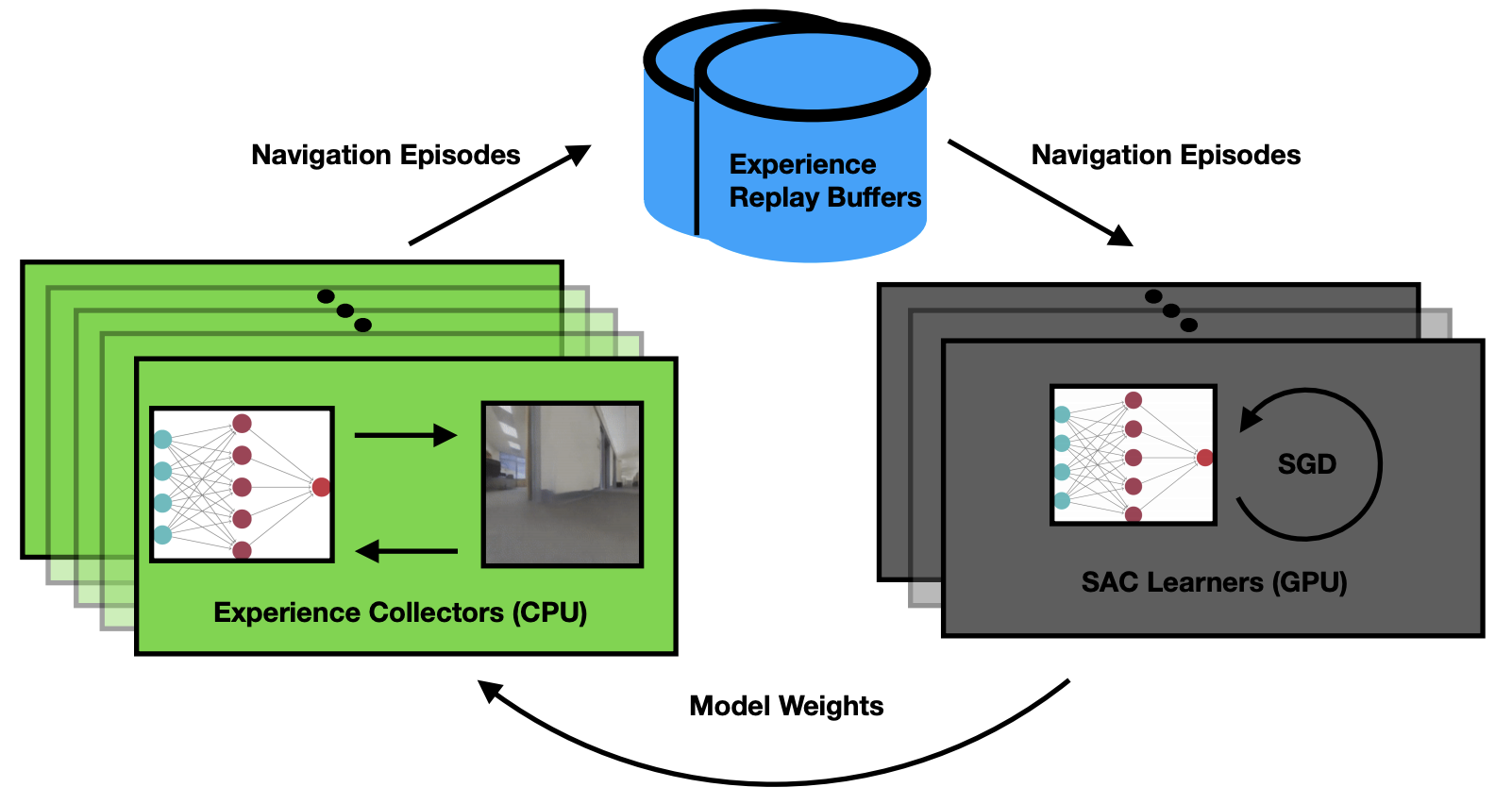}}
        \caption{\label{fig:my-label} Diagram of distributed Soft Actor Critic architecture.}
        \label{fig:policy_architecture}
\end{figure}

As commonplace, we model the problem as a Partially Observable Markov Decision Process (POMDP). In particular, we apply Soft Actor-Critic to learn a policy for the MDP~(see \cite{haarnoja2018soft} and Supplement). As SAC is an off-policy RL algorithm, it lends itself well to a distributed application. SAC has been empirically demonstrated to work well with significantly out-of-distribution data, thus this setup is very amenable to parallelization and distribution. In particular, we can distribute several aspects of the process (see Fig.~\ref{fig:my-label}).

\textbf{Collect} The data collection can be parallelized over multiple workers, on which we run our policy in the simulation engine. In our implementation we use a simulator that renders GibsonEnv worlds~\cite{xia2018gibson} and simulates a LoCoBot differential drive base~\cite{wogerer2012locobot}. Note that most of the running time is spent in stepping through the environment. Many other navigation simulation engines have GPU specific renderers~\cite{xia2018gibson,savva2019habitat} to speed up graphic rendering. In our design we opted for running $100$ CPUs in parallel. 

\textbf{Training} We use a rack of $8$ GPUs to perform synchronous SGD. We use the SAC implementation from the Tensorflow Agents library within TensorFlow~\cite{abadi2016tensorflow}.

\textbf{Experience Replay} To mix experiences from multiple collect workers, we use an Experience Replay Buffer (ERB). The ERB receives experiences generated by all collect jobs and sends them to the training worker. In more detail, each collect worker sends an unroll (of a length of up to $100$ steps in our implementation) to the ERB . The ERB sends batches of samples (of length $20$ each) to the workers. Each sample is generated as a random crop from an unroll. In this way, the ERB guarantees uniform sampling of the data. Note that we use a distributed implementation of ERB due to its large memory requirements.

\textbf{Performance Analysis} The computational bottleneck in the above setup is the training, as the training time scales approximately logarithmically w.~r.~t.~the number of workers~\cite{chen2016revisiting}, while the collect process scales linearly. Therefore, it is important to reduce additional computational burden on the training workers. We do this by moving experience collect on the separate collect workers. In our experiments we can process $98$ million experience steps per $8$ GPUs per day. %For comparison Wijmans et al.~\cite{wijmans2019dd} very recently reports $100$ million experience steps per $8$ GPUs\footnote{The comparison is not apples-to-apples as both this work and \cite{wijmans2019dd} use modified ResNet50 image embedders and LSTMs of different sizes. Further, we use Tesla P100, while \cite{wijmans2019dd} uses the more modern V100.}

\subsection{Reward Definition for Object Navigation}\label{sec:reward}
\vspace{-2mm}

The only aspect specific to object navigation in the above POMDP is the reward definition used at train time. We deliberately opt for a simple reward, which provides clear supervision for clearly desired and undesired behaviors, such as reaching goal or colliding respectively. At the same time, we provide no supervision for behaviors that are hard to quantify, such as efficient exploration behaviors or effective low level controls.

In more detail, the reward consists of four components:
\[
  R(s, a) = \textrm{collision}(s) + \textrm{step}(s) + \textrm{progress}(s, a) + \textrm{success}(s)
\]
where $\textrm{collision}(s) = -0.05$ if $s$ denotes a state of collision, otherwise $0$. $\textrm{step}(s) = -0.01$ which is used to encourage efficiency (penalizes long episodes). The third term measures the distance progress towards the goal, without taking into account rotations. More precisely, if $s'$ denotes the state after taking action $a$ at $s$ and $d(s)$ denotes geodesic distance to goal, then $\textrm{progress}(s, a)=0.1(d(s)-d(s'))$. The final term, $\textrm{success}(s)=1.0$ if state $s$ denotes success of reaching the goal object (see Sec.~\ref{sec:exp_setup} for detailed definition for object navigation). 

\subsection{Observations, Actions, and Network Architecture}\label{Sec:observations}
\vspace{-2mm}
\textbf{Observations} Our agent runs with three different observation modalities: forward facing RGB image, ground level LiDAR, and depth image, as commonly found on robots. The \texttt{RGB} observation is a frontal camera view from the agent's perspective at 0.88m height, perpendicular to the ground plane with a horizontal FOV of 79 degrees. This setup mimics the setup of a LoCoBot with an elevated Intel RealSense Camera. It gives the agent the opportunity to see the world directly and use semantic cues. However, the camera at this height has a blind spot immediately in front of the robot base, which makes obstacle avoidance challenging. In order to perceive geometry, we use a depth image called \texttt{DEPTH} of the same size and extrinsics as \texttt{RGB}. 

Lastly, the \texttt{LiDAR} observation is a 1-dimensional array of 222 agent-relative (x, y, z) Cartesian coordinates, which indicate the ray-distance to occupied space in the environment. It is mounted at the base of the robot, has a wider field of view than \texttt{DEPTH} at 220 degrees, and is motivated by the blind spot of \texttt{RGB}. It is used by real world robotic systems such as Fetch, and can be modelled on a LoCoBot by using a base mounted Hokyuo Rangefinder.

%These observations are visualized in \ref{fig:fig7}. (TODO: ayzaan, check the following) The RGB image, depth image, and segmentation all concatenated along the channels axis and are processed by a resnet based convolutional network. The LIDAR observation array is then concatenated with a feature vector extracted from the image modalities and fed into the recurrent portion of our architecture.

\textbf{Auxiliary Observations} In addition to the above observations, we feed into our policy two other auxiliary observations shown to benefit learning: the action from the previous time step, a binary collision bit designating whether the action at the previous step was successful. These observations help the agent learn collision avoidance, and empirically leads to faster convergence.

\textbf{Oracle Observations}
We conduct experiments with auxiliary oracle observation to understand the importance of semantics. This oracle observation is defined as a binary mask of only the goal object, called \texttt{Det}. This observation helps break the problem into two distinct phases: exploration (when the object is not visible) and direct navigation to the object while avoiding obstacles (once it becomes visible).

%Finally, we conduct experiments by feeding the pose of the agent w.~r.~t.~the closest goal object and refer to this observation as \texttt{GoalVector}. It ablates against the need of exploration even when the goal is not visible.

\textbf{Action Space} The action space of the agent is motivated by commonly used differential drive robots such as LoCoBot or Fetch. We use twist commands, an angular and linear velocity.

\textbf{Architecture} We use a standard NN architecture, whereby the different observations get embedded into vectors. These get concatenated, and subsequently fed into an LSTM, which outputs actions and critic values. Further details regarding the exact architecture and training details can be found in supplementary material. 

\section{Empirical Analysis}\label{sec:exp_setup}
\vspace{-2mm}
\subsection{Setup}
\vspace{-2mm}
We evaluate our system in an environment which simulates both the visuals and physics of the real world. We use the worlds from Gibson~\cite{xia2018gibson}, which consists of high quality scans of real homes, and simulate a differential drive robot in a custom environment for performance reasons. We use a subset of the worlds for which object segmentations and labels are available, as provided in \cite{armeni20193d}. This dataset comes with $25$ homes for training, $5$ for validation, and $5$ for testing. Note that many of the homes have multiple floors, which we treat as disjoint spaces. For example, the test set has $12$ floors total. We test navigating to 9 objects: `bed', `chair', `microwave', `refrigerator', `table', `toilet', `oven', `tv', 'sofa'.

We model the kinematics of a commonly used differential drive robot, i.e.,~LoCoBot or Fetch. Its base is a $0.18$m radius circle and its camera height is $0.88$m. To evaluate each approach we compute Success weighted by Path Length (SPL)~\cite{anderson2018evaluation}.
% This metric, defined over $N$ navigation episodes, is the average of the success indicators $S_i$  scaled by navigation efficiency, expressed in terms of the optimal path length $o_i$ to goal and the length $l_i$ of the actual path taken: $\textrm{SPL}=\frac{1}{N} \sum_{i=1}^N S_i \frac{o_i}{\max\{l_i, o_i\}}$. We note that a perfect SPL of 1.0 is not realistic to expect because navigation decisions in unexplored environments can be ambiguous.

\begin{figure}[t]
        \centering\includegraphics[width=0.45\textwidth]{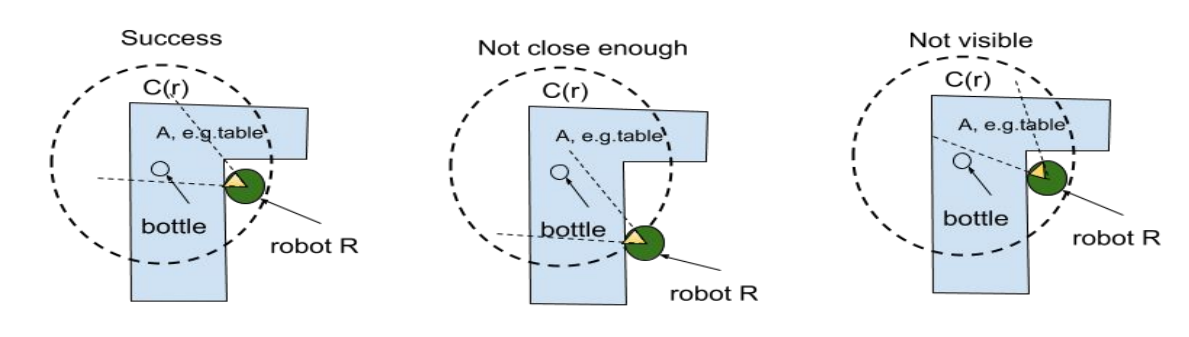}
        \includegraphics[width=0.45\textwidth]{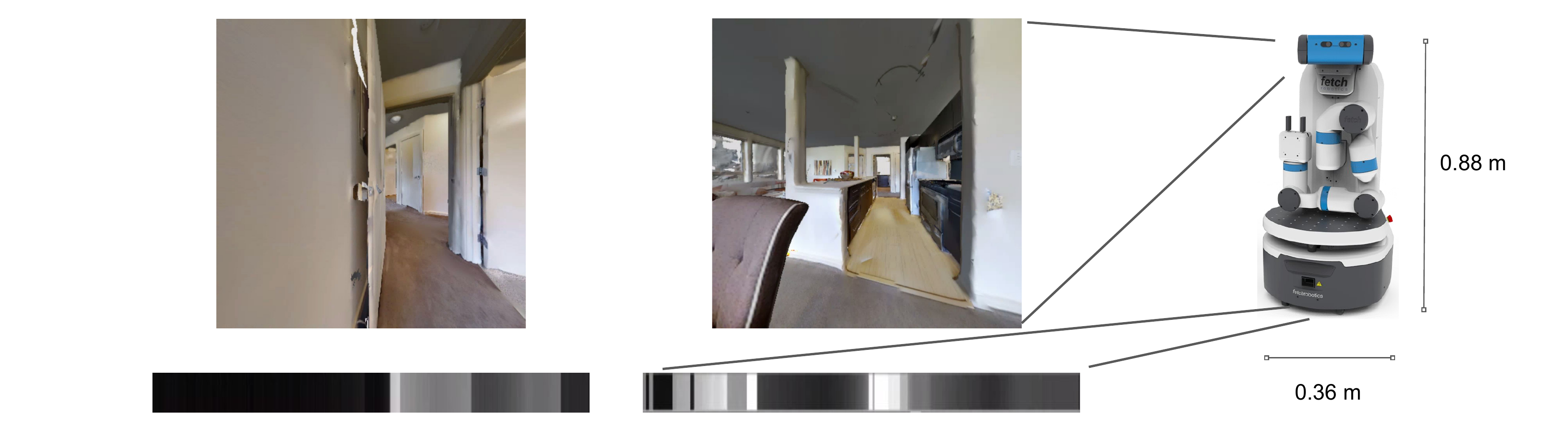}
        \caption{\label{fig:objnav_term} Left: A robot is considered to have reached the target bottle placed on corner desk if it is looking at the bottle and is within $1m$ from bottle surface. Right: \texttt{RGB} (top) and \texttt{LiDAR} observations (bottom) from the environment. Note that the camera is at a height of $0.88$m with $79$ degrees FOV which makes collision detection challenging. Using range sensing close to the base makes the control problem easier, e.~g.~dark means close and it demonstrates that the robot is quite close to the wall in the left image and the chair in the right image.}
% \caption{\label{fig:observation_examples} \texttt{RGB} (top) and \texttt{LiDAR} observations (bottom) from the environment. Note that the camera is at a height of $0.88$m with $79$ degrees FOV which makes collision detection challenging. Using range sensing close to the base makes the control problem easier, e.~g.~dark means close and it demonstrates that the robot is quite close to the wall in the left image and the chair in the right image.}
\end{figure}

The success criterion, needed to compute $S_i$, is motivated by the idea that the robot should be able to physically reach the object. In particular, we expect that the agent is looking at the object and that the object is within reaching distance (see Fig.~\ref{fig:objnav_term}, Left). More precisely, $S_i=1$ if agent is at a navigable pose within $1m$ from the object 3D bounding box and the object is visible (within the central 20\% of the \texttt{RGB} observation), otherwise $S_i=0$.

% In addition, we would like to quantify the performance of the low-level control. For this, we measure the fraction of steps resulting in collision, called Collision Free Rate (CFR). More precisely, $\textrm{CFR} = \frac{1}{N} \sum_{i=1}^N \frac{f_i}{l_i}$ where $f_i$ is the number of collision free steps in the $i^\textrm{th}$ episode.

\subsection{Experiments}
\vspace{-2mm}
In this section we present an empirical analysis of the learned policy. To focus on understanding exploration and low-level control, we assume that we are given an oracle goal object detection in most of the analysis below (see Sec.~\ref{Sec:observations}). 

We report results over $120$ episodes, $10$ episodes for each of the $12$ house floors. Each episode runs for up to $500$ steps. The starting position is spawned randomly within $15$ m of the goal. Given that we use scans of standard homes, this distance is large enough that the goal object is usually not visible from the starting point.

Using modalities \texttt{RGB}+\texttt{LiDAR}+\texttt{Det} our navigation policy achieves an SPL of $0.58$ and success rate of $0.68$ on Gibson Test environments (see Table~\ref{tab:overview}, left). %The discrepancy between success rate and SPL indicates that the policy has to explore the physical space, as in successful cases it does not take the shortest path, i.~e.~it needs to explore.
% First, contrary to many supervised problems, where there is a clear label inferable by a human, this isn't the case for object-conditioned exploration. Humans do not have always the needed the knowledge to take the shortest path to the goal.
In order to provide context to this result, we compare with several baselines. Because the environment is only partially observable, it is not reasonable for any agent to achieve 1.0 SPL. Thus, to get a sensible upper bound on performance, we ask $12$ human raters to perform navigation using $\texttt{RGB}$ observation by virtually traversing 12 environments (see appendix for details). The average SPL for our raters is $0.80$ with a success rate of $0.86$. The challenges raters face are: not being able to identify the correct direction, which leads to a suboptimal path; running out of time; getting stuck in narrow spaces, where the head camera view is not allowing for collision free traversals.

We compare  against two scripted baselines with varying degrees of exploration intelligence to quantify how much learning exploration helps. For all of these baselines, the agent goes straight and collision-free to the goal whenever the object becomes visible. 

\textbf{Roomba:} the policy moves straight until collision. When it collides, it picks a random rotation direction, performs rotations until collision free, and repeats.

\textbf{Topo Graph Traversal (TGT):} the policy follows a topological graph of the environment, defined as the skeleton of the floorplan~\cite{kuipers1988navigation}. We perform a depth-first traversal of this graph. This guarantees an efficient coverage of the space, however the space is explored without taking into account the goal label. This baseline is by definition collision free.

%As the problem of object navigation has three aspects: recognition, exploration, and low-level control, we introduce oracle baselines for which there is a solution for some of these aspects. In all cases, we assume we have perfect low-level control, i.~e.~the robot is controlled using position control (no need to output twist commands) without any collisions. 

%The DFT baseline is by design collision-free navigation. Further, it uses an oracle traversal mechanism whereby the free space is efficiently covered by performing a Depth First traversal of the topological graph of the space. This traversal assures that the space is covered in an efficient manner. Under unlimited budget, it should eventually navigate to the object, However, no contextual guidance is being exercised, e.~g.~if at an intersection, the robot is to take the next branch of the topological graph without leveraging any visual cues. 

Both scripted baselines achieve similar SPL, however Roomba has a lower success rate as it struggles exploring far-away spaces. TGT can do this, however often times the path taken is not very efficient. Our navigation policy achieves almost double the SPL. Note that humans are not perfect either.

%In particular, when starting more than $7$ meters away from the target DFT achieved $0.0$ SPL, while SACNav achieves SPL of $0.38$ when starting between $7$ and $10$ meters away, and SPL of $0.10$ when starting between $10$ and $15$ meters away. This shows that the learned guided exploration executed by SACNav is crucial in getting to the target in a limited number of steps.

\begin{table}
\begin{center}
\begin{tabular}{|l|c|c|}
\hline
  & SPL & SR \\ 
\hline\hline
 Human Navigator & 0.80 & 0.86 \\  
\hline
 \texttt{RGB} + \texttt{LiDAR} + \texttt{Det} & 0.58 & 0.68 \\
 \hline
 TGT& 0.32 & 0.50 \\
\hline
 Roomba & 0.30 & 0.35 \\
 \hline
\end{tabular}\quad\quad
\begin{tabular}{|l|c|c|c|c|}
\hline
  \multirow{2}{*}{Modalities} &
  \multicolumn{2}{c|}{\tiny stuck at collision} &
  \multicolumn{2}{c|}{\tiny sliding at collision} \\
 \cline{2-5}
 & SPL & SR & SPL & SR \\ 
\hline\hline
 \texttt{RGB} + \texttt{Det} & 0.14 & 0.14 & 0.28& 0.30\\  
\hline
 \texttt{Depth} + \texttt{Det} & 0.22 & 0.25 & 0.39& 0.45\\
\hline
 \texttt{RGB} + \texttt{Depth} + \texttt{Det} & 0.26 & 0.30 & 0.46 & 0.55\\
 \hline
 \texttt{RGB} + \texttt{LiDAR} + \texttt{Det} & 0.58 & 0.68 &0.68 & 0.79\\
 \hline 
 \texttt{RGBD} + \texttt{LiDAR} + \texttt{Det} & 0.52 & 0.66 & 0.63&0.75 \\
 \hline
%  \hline
%  \texttt{RGB} + \texttt{Depth} + \texttt{LiDAR} &0.28 &0.35 \\
%  \hline
\end{tabular}
\end{center}
\caption{\label{tab:overview}Success weighted by Path Length (SPL) and object reached Success Rate(SR). Left: comparison of our approach with scripted and human baselines. Right: different combinations of observation modalities. We have two behaviors of the simulator: `stuck', proper physics dynamics whereby the agents usually gets stuck at collision, and `sliding', where the agent slides upon collision as used in \cite{savva2019habitat}.}
\vspace*{-1.0cm}
\end{table}
\textbf{Observation Modalities}
Oftentimes it is assumed that \texttt{RGB} and \texttt{Depth} are sufficient observation modalities. However, on real world robotic systems a stronger emphasis is put on range sensing or wider field of view as they usually give complementary information. For example, when mounted on top of a robot, both \texttt{RGB} and \texttt{Depth} have a blind spot right in front of the robot (see Fig.~\ref{fig:objnav_term}, Right). This blind spot makes it harder to maintain a collision free path. Therefore, in this work we mount a \texttt{LiDAR} sensor to the robot base and investigate its utility.

We present results over combinations of modalities in Table~\ref{tab:overview}, right. In all cases we use \texttt{Det}, so that the policy can focus on exploration and continuous control. We see that adding depth sensing leads to a boost, and in particular \texttt{LiDAR} has the larger boost in performance. We conjecture that this is due to the ability to avoid collisions, in which the agent otherwise gets stuck.

\textbf{Low level continuous control} We use a simulator for our differential drive robot model. Empirically, the most challenging aspect in that respect is collisions, especially in narrow passages, which lead to a halt.

In order to assess how much of challenge this presents, we evaluate our policies in a modified version of our simulator where, at collision, the agent is allowed to slide along the obstacle, as used in~\cite{savva2019habitat}. The results, shown in Table~\ref{tab:overview}, indicate a $0.10$ to $0.20$ boost in SPL across all policies in this more forgiving regime. This suggests that properly modeling dynamics, especially in the physically realistic continuous control case, is of high importance. 

% \begin{figure}[t]
% \centering
%     \includegraphics[width=0.75\textwidth]{figs/spl_ranges2.pdf}
%     \caption{\label{fig:spl_ranges}SPL for different starting distances to goal and all modality combinations. }
% \end{figure}
\textbf{Starting Distance to Goal}
As the starting distance to the goal increases, the problems become harder (see Fig.~\ref{fig:spl_ranges}). It is noteworthy that \texttt{LiDAR}-based policies tend to perform better over larger distances, which is due to better collision avoidance (usually narrow doorways in our data).

\textbf{Visual Recognition}
In most of the experiments we assume that an oracle gives us goal object recognition via \texttt{Det}, in order to focus on exploration. To demonstrate that our system is capable of learning recognition in addition to exploration and continuous control, we train our policy without \texttt{Det} (see Table~\ref{tab:spl_per_object}). \texttt{Det} helps across the board, in particular when the object is far away and small as a result.
% \begin{table}
% \begin{center}
% \begin{tabular}{|l|c|c|c|c|c|c|c|}
% \hline
%   & bed & chair & oven & fridge & sofa & toilet &  mean \\ 
% \hline\hline
% \texttt{RGB}+\texttt{Depth}+\texttt{LiDAR}   & 0.40 & 0.41 &  0.41 &  0.65 &  0.36 &  0.04 & 0.28\\
% \hline
% \texttt{RGB}+\texttt{Depth}+\texttt{LiDAR}+\texttt{Det}  & 0.73 & 0.53 & 0.39 & 0.48 & 0.77 & 0.36& 0.52 \\ 
%  \hline
% \end{tabular}
% \end{center}
% \caption{\label{tab:spl_per_object}SPL for our model with and without \texttt{Det}, also split per object.}
% \vspace{-0.8cm}
% \end{table}

\begin{figure}
\centering 
% 4x1 figure widths: .202, .342, .23, .202
% 2x2 figure widths: .365, .62, .53, .46
\begin{subfigure}{.202\textwidth}
\includegraphics[width=\textwidth]{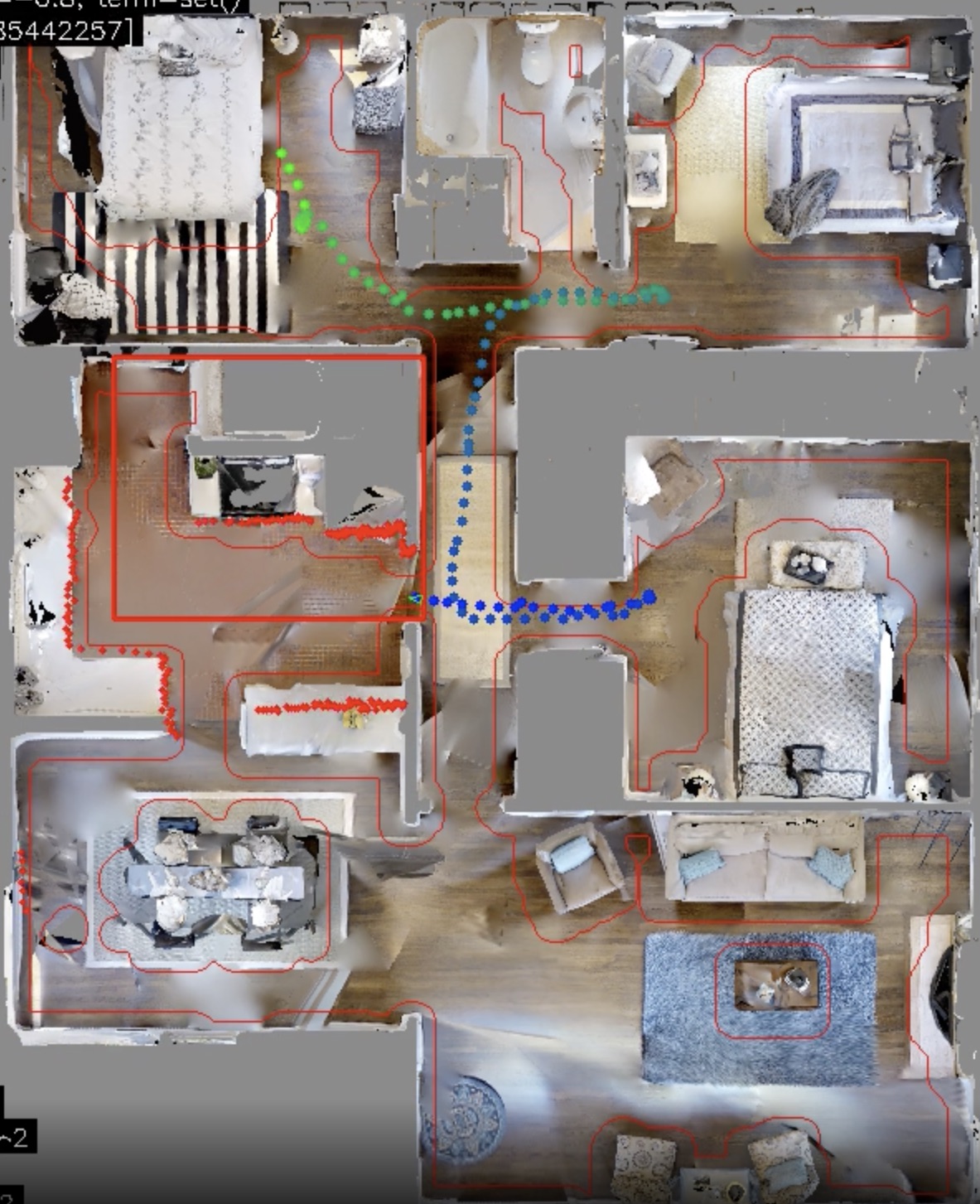}
\caption{Peeking}\label{fig:efficient}
\end{subfigure}
\begin{subfigure}{.342\textwidth}
\includegraphics[width=\textwidth]{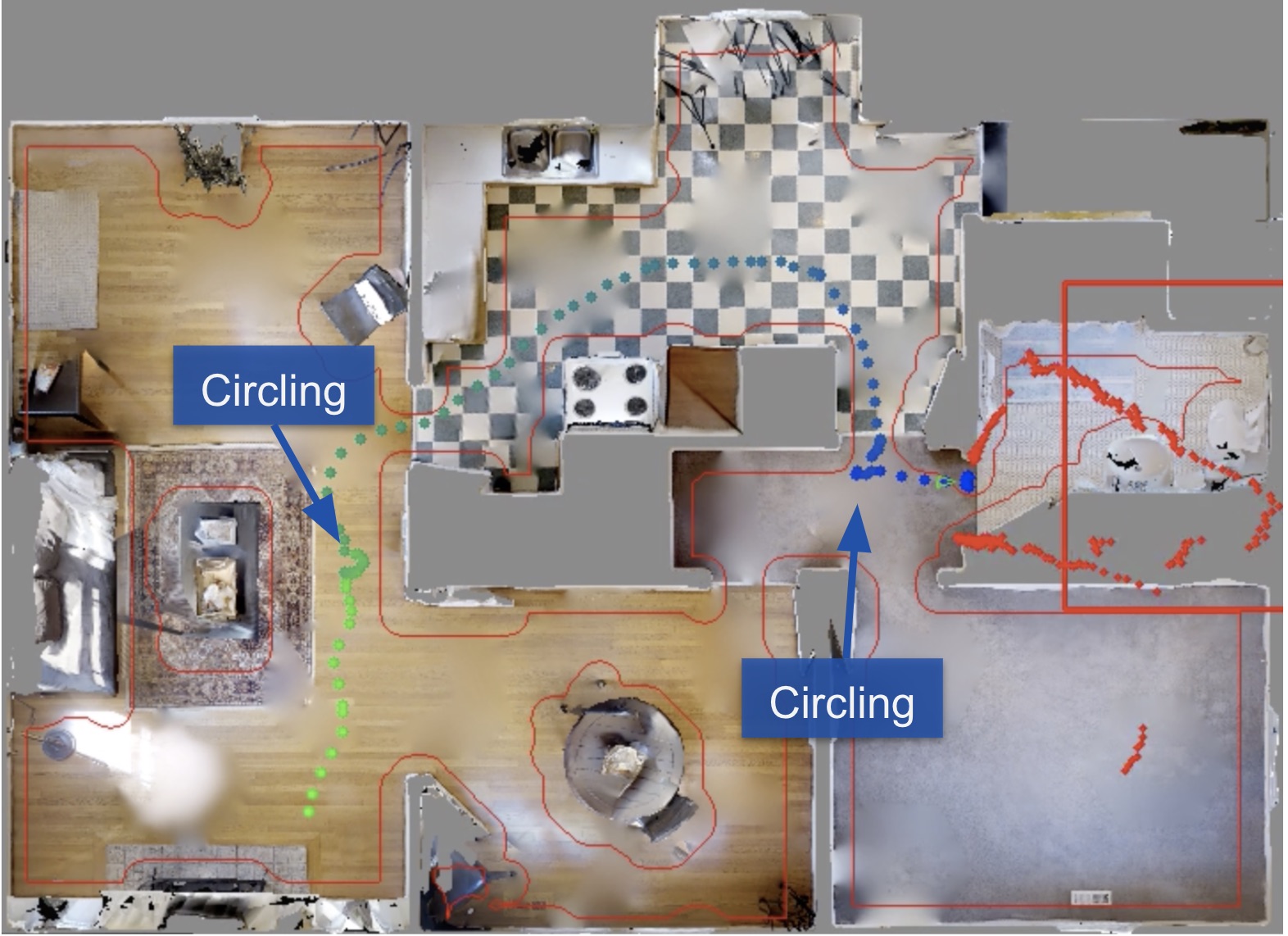}
\caption{Circling}\label{fig:circling_labeled}
\end{subfigure}
\begin{subfigure}{.23\textwidth}
\includegraphics[width=\textwidth]{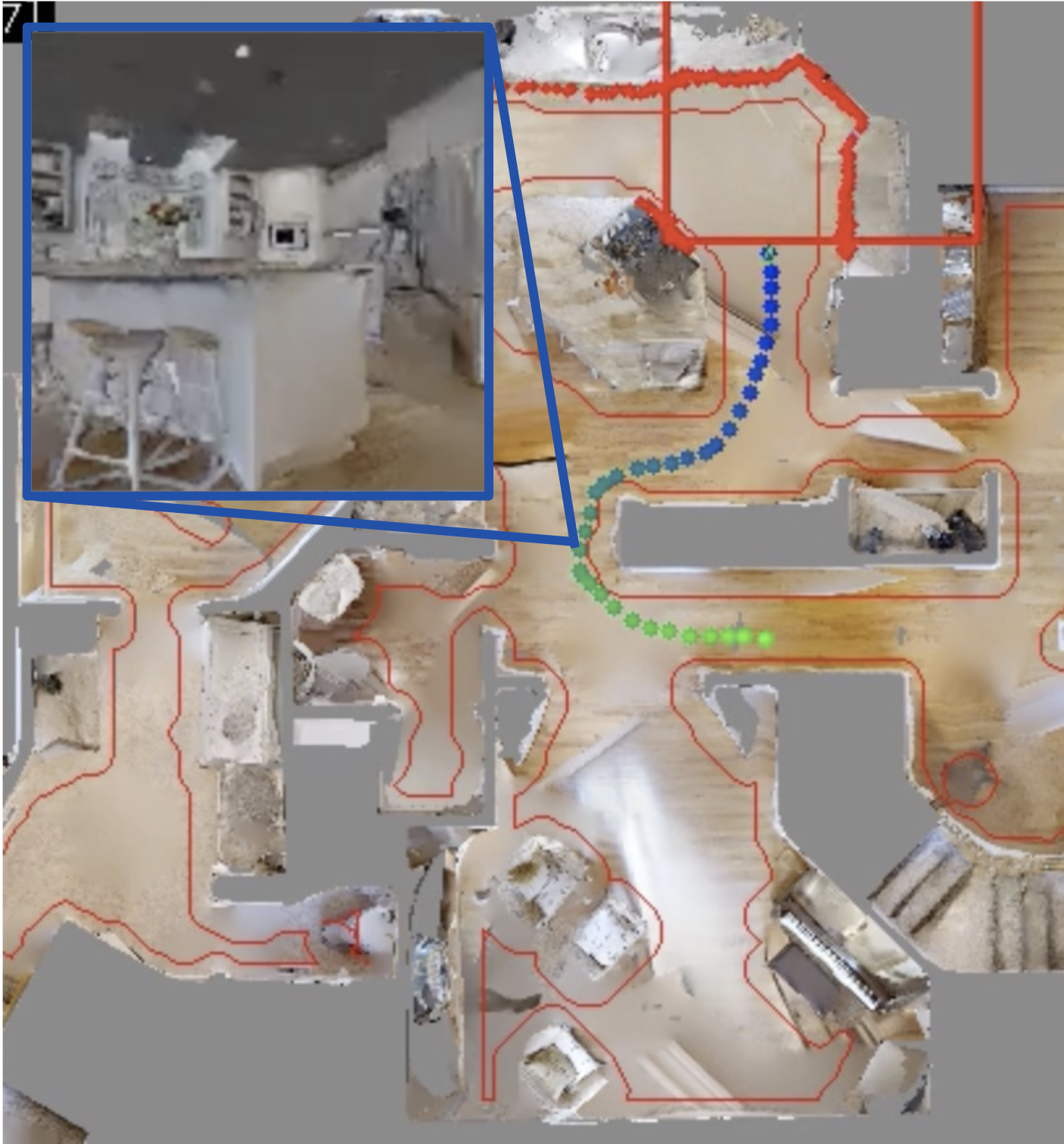}
\caption{Context}\label{fig:context}
\end{subfigure}
\begin{subfigure}{.202\textwidth}
\includegraphics[width=\textwidth]{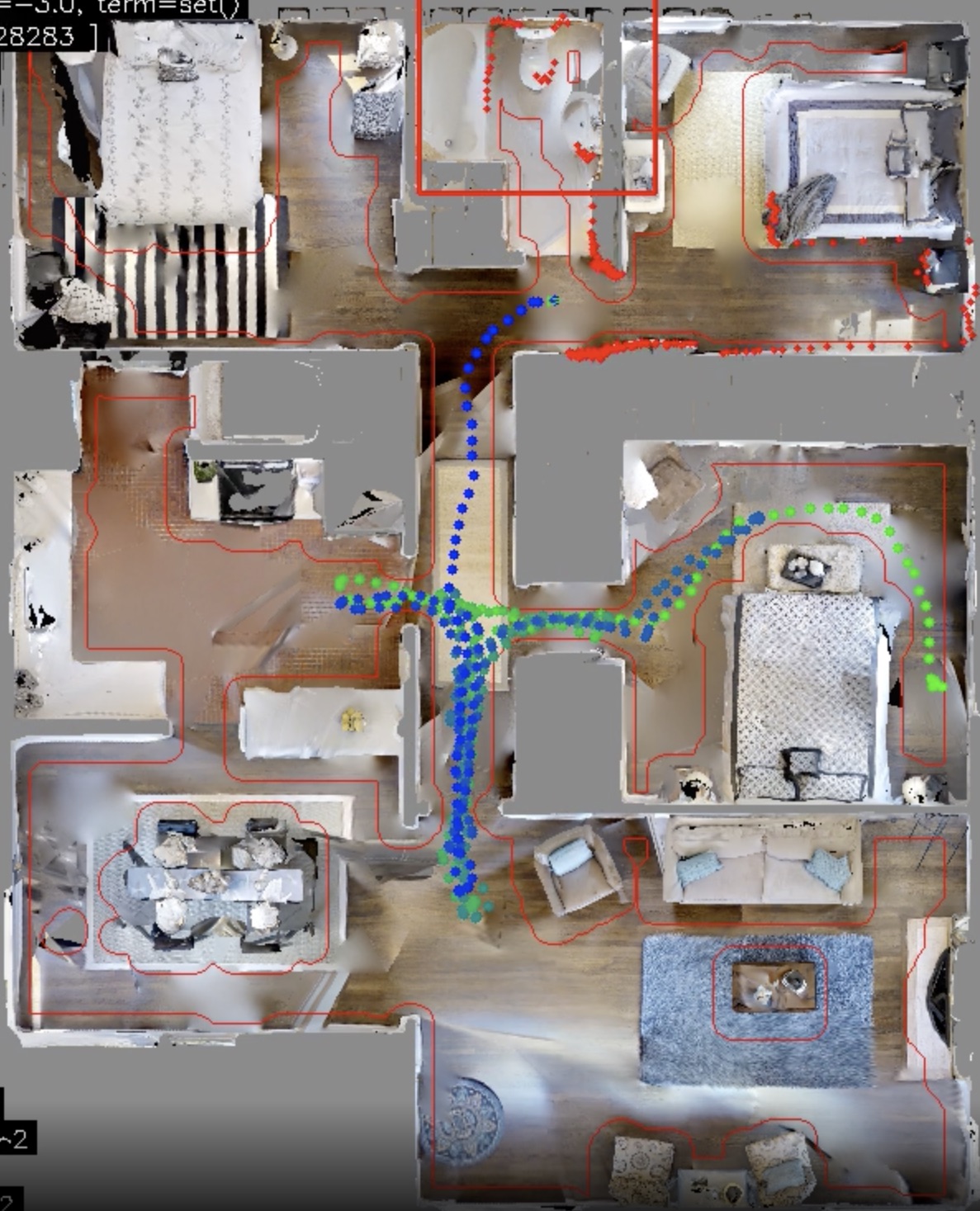}
\caption{Fail: Revisits}\label{fig:failure_doubling}
\end{subfigure}
\caption{For each figure, the trajectory is colored from green to blue (start to end), and the goal location(s) are indicated via a red box.
(\ref{fig:efficient}) The agent efficiently explores the environment, exploring rooms only the amount necessary to confirm the lack of the the target object.
(\ref{fig:circling_labeled}) The agent alternates between directly moving through the environment, followed by circling behavior to gather information. 
(\ref{fig:context}) The agent leverages context of the scene. Even without seeing the oven, the agent recognizes the kitchen and approaches the area. 
(\ref{fig:failure_doubling}) At times the agent doubles back and revisits previously explored regions, indicating a need for longer-term memory.}
\label{fig:behaviors_exp}
\end{figure}

\begin{table}[ht]
\begin{minipage}[b]{0.45\linewidth}
\centering
    \includegraphics[width=\textwidth]{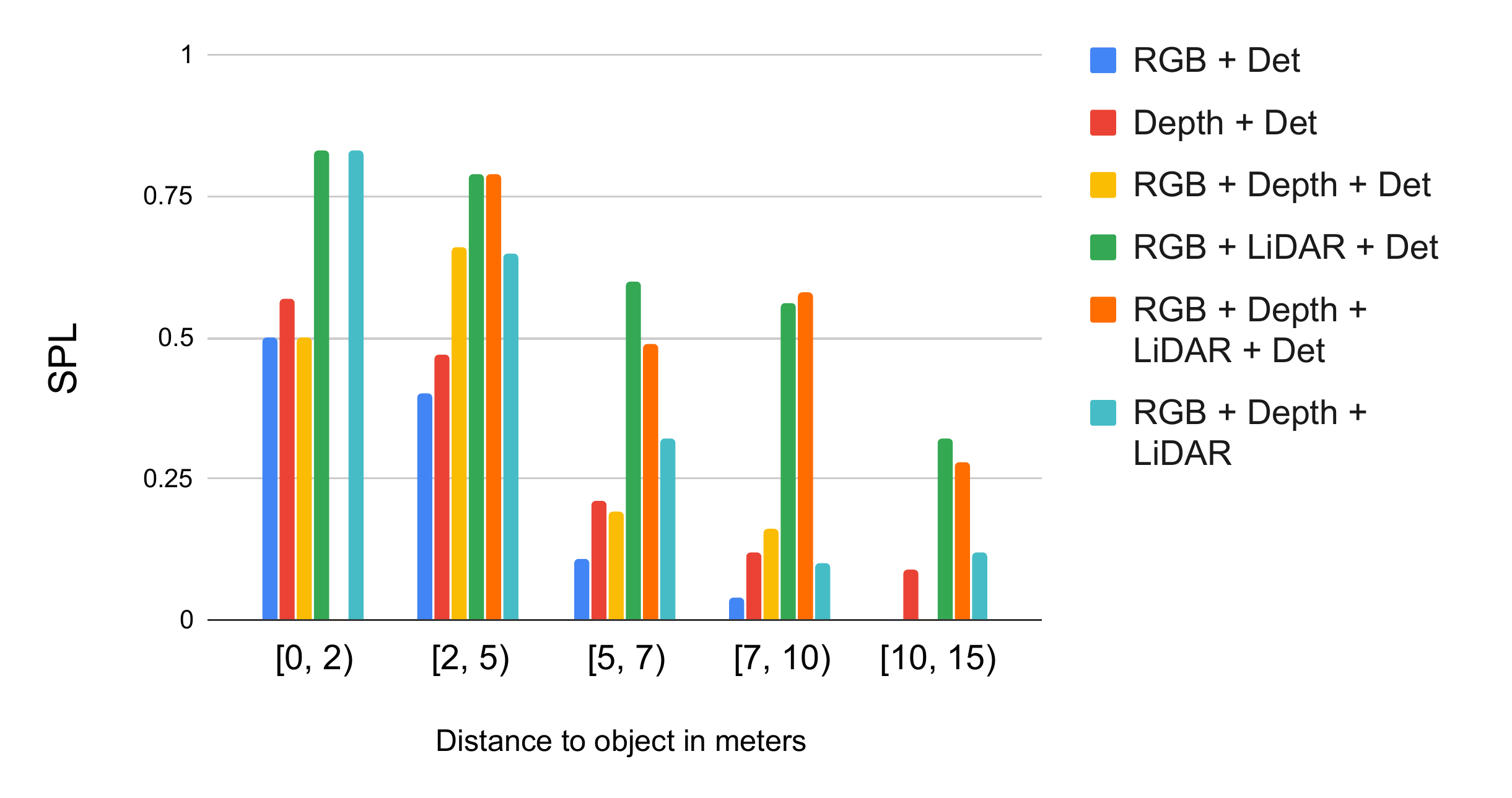}
    \captionof{figure}{\label{fig:spl_ranges}SPL for different starting distances to goal and all modality combinations. }
\end{minipage}
\hfill
\begin{minipage}[b]{0.45\linewidth}
\centering
\begin{tabular}{|l|c|c|}
\hline
object & \texttt{RGBDL} & \texttt{RGBDLDet} \\
\hline
\hline
bed & 0.40 & 0.73 \\
\hline
chair & 0.41 & 0.53 \\
\hline
oven & 0.41 & 0.39 \\
\hline
fridge & 0.65 & 0.48 \\
\hline
sofa & 0.36 & 0.77 \\
\hline
toilet & 0.04 & 0.28 \\
\hline
\hline
mean & 0.28 & 0.52 \\
\hline
\end{tabular}
\caption{\label{tab:spl_per_object}SPL for our model with and without \texttt{Det}, also split per object.}
\end{minipage}
\end{table}
\textbf{Emergent Behaviors}
To solve object navigation, several exploration and navigation behaviors must emerge. 
Towards exploration, an agent must be able to gather information about the environment efficiently and comprehensively.
We note that several of these behaviors naturally emerge through the learning process, as shown in Fig.~\ref{fig:behaviors_exp}.
The agent learns not only to only cover the space, but to leverage the minimal required amount of exploration, e.g., the agent \textit{peeks} into rooms only the necessary amount to see if the object is present.
The agent also learns to efficiently gather information in regions that must be traversed, e.g., by repeatedly moving in straight lines, followed by \textit{circling} to get a 360$^\circ$ view.
Finally, the agent learns to leverage semantic \textit{context} of the surrounding scene to make decisions even when the object is not in view, e.g., recognizing that a kitchen likely has an oven or that a bathroom has a toilet.
We do however find suboptimal behaviors emerging, such as \textit{revisiting} regions or heavily exploring single regions. This is likely a result of the limited memory of LSTMs and indicates a need for further research into long-term memory for navigation.

\begin{figure}
\centering
\begin{minipage}{0.9\textwidth}
\begin{subfigure}{.487\textwidth}
\includegraphics[width=\textwidth]{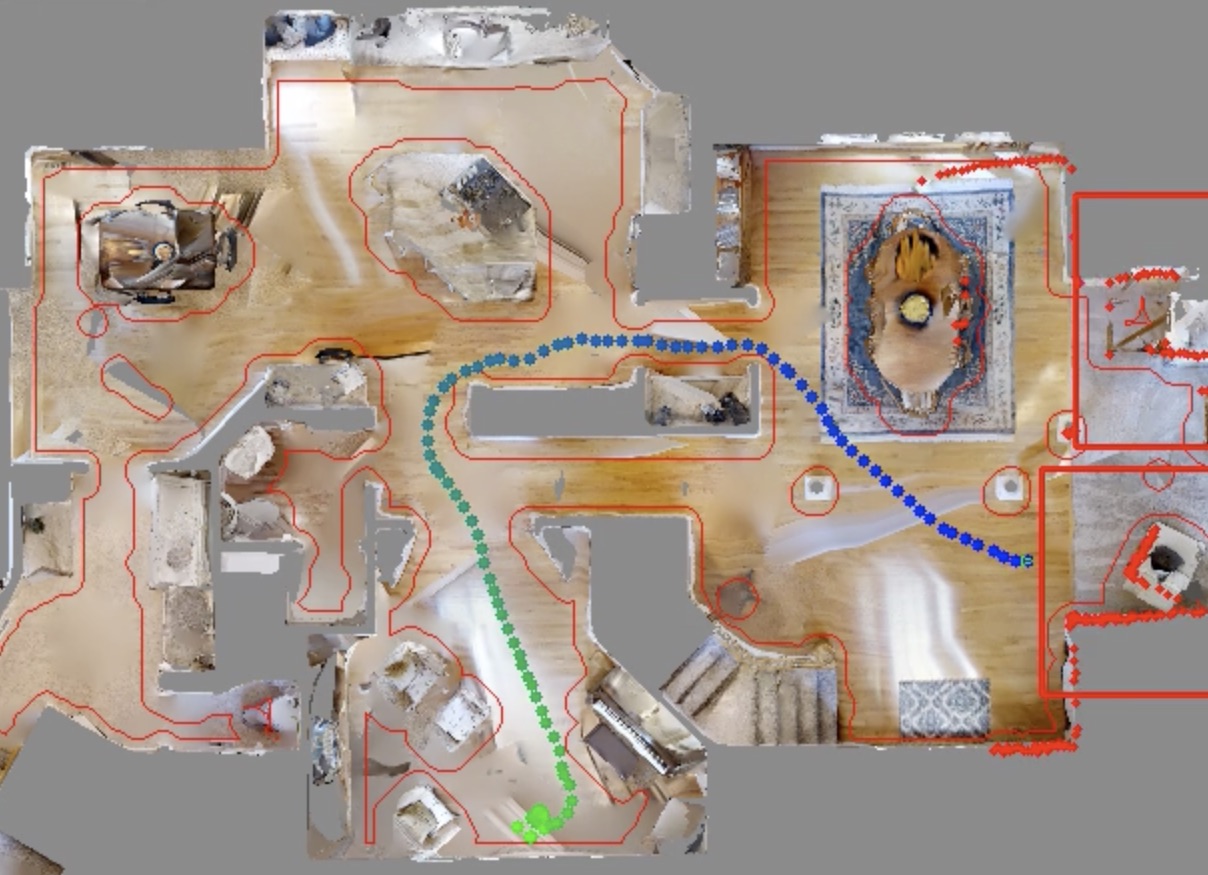}
\caption{Beelining}\label{fig:move01}
\end{subfigure}
\begin{subfigure}{.285\textwidth}
\includegraphics[width=\textwidth]{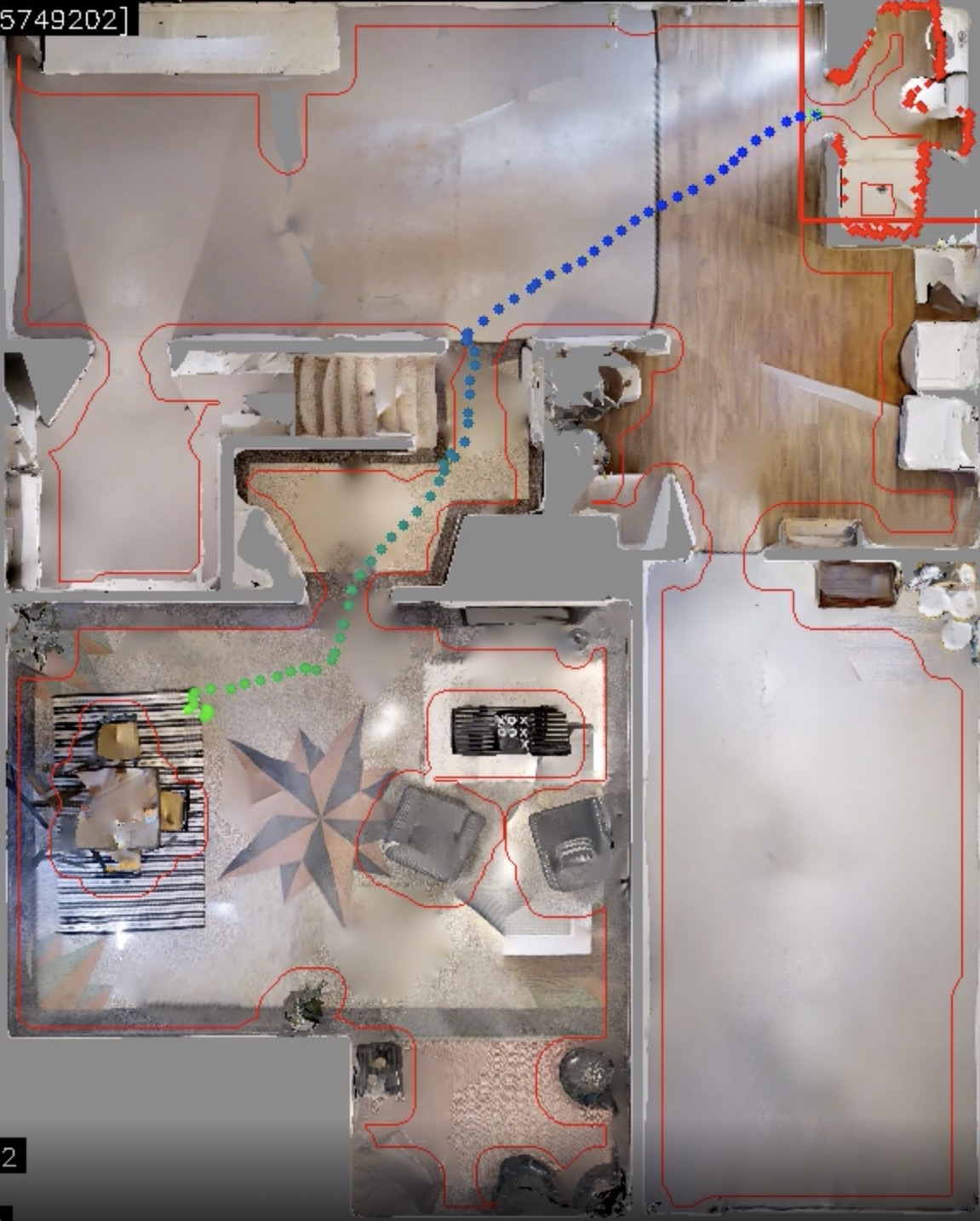}
\caption{Narrow Passages}\label{fig:move02}
\end{subfigure}
\begin{subfigure}{.211\textwidth}
\includegraphics[width=\textwidth]{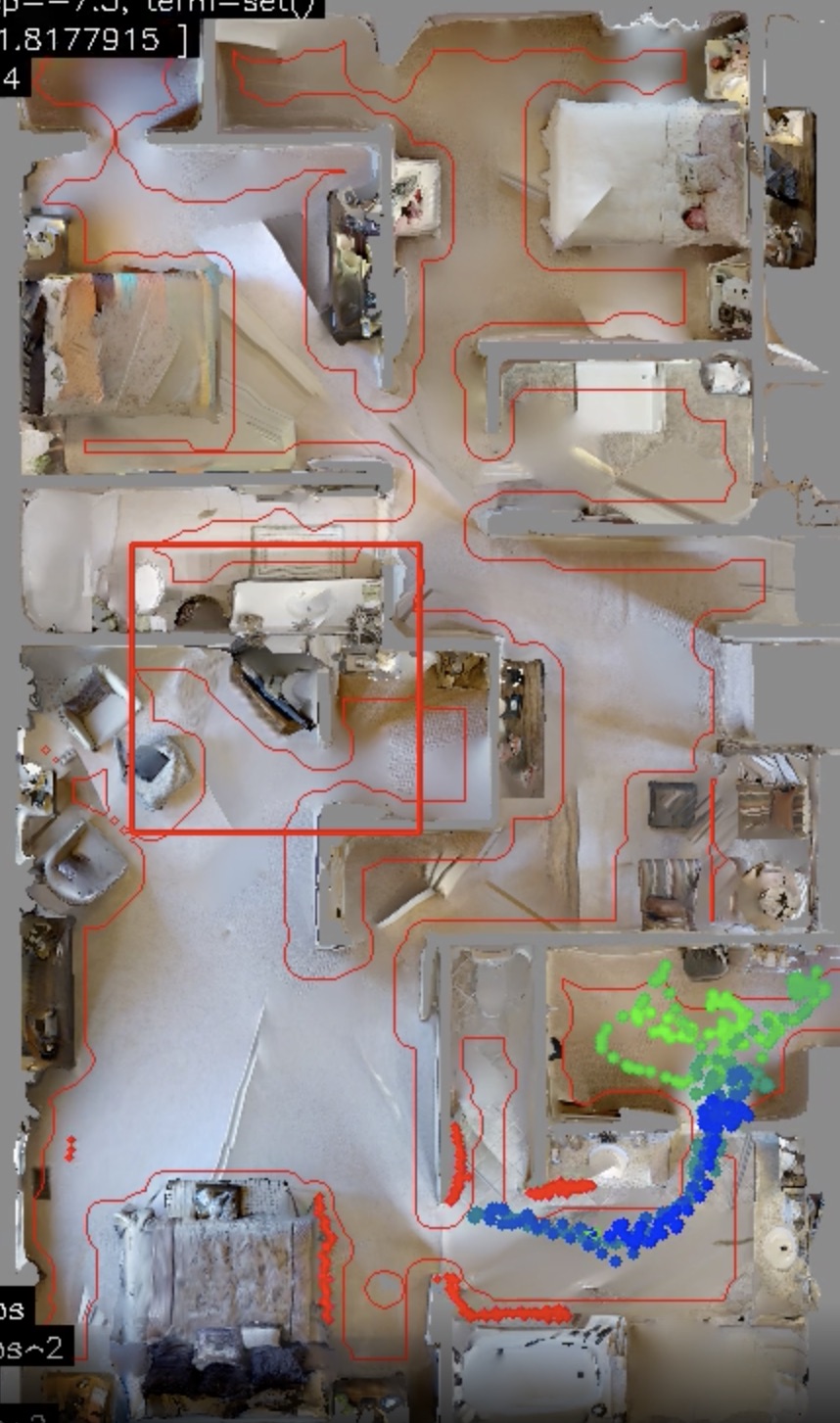}
\caption{Fail: Stuck}\label{fig:failure_stuck}
\end{subfigure}
\end{minipage}
\caption{For each figure, the trajectory is colored from green to blue (start to end), and the goal location(s) are indicated via a red box.
(\ref{fig:move01}) The agent first establishes itself through circling and then moves through the environment in direct, straight lines. Furthermore, once the object is identified, the agent takes the most direct route.
(\ref{fig:move02}) The agent is able to navigate efficiently through narrow passages. 
(\ref{fig:failure_stuck}) The agent occasionally gets stuck within regions of the map or in collision with objects outside its field of view; difficult challenges for embodied agents.}
\label{fig:behaviors_nav}
\vspace{-3mm}
\end{figure}

Towards navigation, several challenging problems and emergent behaviors become apparent when considering a formulation of the navigation problem with continuous control and an embodied agent (Fig.~\ref{fig:behaviors_nav}).
To maximize its ability to both cover the space and to reach known goals, the agent learns to move with direct, smooth trajectories.
This behavior is particularly visible once the object has been found, in which case, the agent \textit{beelines} to the goal.
The agent also learns to identify and traverse \textit{narrow passages}, e.g., doors. 
We find that while the agent is capable of entering doorways, it at times gets \textit{trapped} in regions either by never deciding to exit a region or getting stuck on objects that are outside its field of view.
These challenges exist particularly for domains like robotics, with embodied agents and continuous control, and warrant further research.

% \subsubsection{Human Navigation Behavior and Results}
% \subsubsection{Importance of Low-level/ Control}
% \subsubsection{Affect of Embodiment}
\vspace{-4mm}
\section{Conclusion}
\vspace{-3mm}
In this paper we study object-conditioned exploration and low-level continuous control in the context of object navigation. We present a generic and scalable RL setup which trains navigation policies with emerging natural exploration behaviors. Further, we show challenges with low-level continuous control, and demonstrate how to address them.

\bibliography{egbib}  % .bib

\begin{thebibliography}{35}
\providecommand{\natexlab}[1]{#1}
\providecommand{\url}[1]{\texttt{#1}}
\expandafter\ifx\csname urlstyle\endcsname\relax
  \providecommand{\doi}[1]{doi: #1}\else
  \providecommand{\doi}{doi: \begingroup \urlstyle{rm}\Url}\fi

\bibitem[Xia et~al.(2018)Xia, Zamir, He, Sax, Malik, and
  Savarese]{xia2018gibson}
F.~Xia, A.~R. Zamir, Z.~He, A.~Sax, J.~Malik, and S.~Savarese.
\newblock Gibson env: Real-world perception for embodied agents.
\newblock In \emph{Proceedings of the IEEE Conference on Computer Vision and
  Pattern Recognition}, pages 9068--9079, 2018.

\bibitem[Haarnoja et~al.(2018)Haarnoja, Zhou, Hartikainen, Tucker, Ha, Tan,
  Kumar, Zhu, Gupta, Abbeel, et~al.]{haarnoja2018soft_constrained}
T.~Haarnoja, A.~Zhou, K.~Hartikainen, G.~Tucker, S.~Ha, J.~Tan, V.~Kumar,
  H.~Zhu, A.~Gupta, P.~Abbeel, et~al.
\newblock Soft actor-critic algorithms and applications.
\newblock \emph{arXiv preprint arXiv:1812.05905}, 2018.

\bibitem[Cadena et~al.(2016)Cadena, Carlone, Carrillo, Latif, Scaramuzza,
  Neira, Reid, and Leonard]{SLAM16}
C.~Cadena, L.~Carlone, H.~Carrillo, Y.~Latif, D.~Scaramuzza, J.~Neira, I.~D.
  Reid, and J.~J. Leonard.
\newblock Simultaneous localization and mapping: Present, future, and the
  robust-perception age.
\newblock \emph{IEEE Transactions on Robotics}, 2016.

\bibitem[Armeni et~al.(2016)Armeni, Sener, Zamir, Jiang, Brilakis, Fischer, and
  Savarese]{armeni_cvpr16}
I.~Armeni, O.~Sener, A.~R. Zamir, H.~Jiang, I.~Brilakis, M.~Fischer, and
  S.~Savarese.
\newblock {3D} semantic parsing of large-scale indoor spaces.
\newblock In \emph{CVPR}, 2016.

\bibitem[Mirowski et~al.(2017)Mirowski, Pascanu, Viola, Soyer, Ballard, Banino,
  Denil, Goroshin, Sifre, Kavukcuoglu, Kumaran, and Hadsell]{Learning2Navigate}
P.~Mirowski, R.~Pascanu, F.~Viola, H.~Soyer, A.~J. Ballard, A.~Banino,
  M.~Denil, R.~Goroshin, L.~Sifre, K.~Kavukcuoglu, D.~Kumaran, and R.~Hadsell.
\newblock Learning to navigate in complex environments.
\newblock \emph{ArXiv arXiv:1611.03673}, 2017.

\bibitem[Gupta et~al.(2017)Gupta, Davidson, Levine, Sukthankar, and
  Malik]{gupta2017cognitive}
S.~Gupta, J.~Davidson, S.~Levine, R.~Sukthankar, and J.~Malik.
\newblock Cognitive mapping and planning for visual navigation.
\newblock In \emph{Proceedings of the IEEE Conference on Computer Vision and
  Pattern Recognition}, pages 2616--2625, 2017.

\bibitem[Zhu et~al.(2017)Zhu, Mottaghi, Kolve, Lim, Gupta, Fei-Fei, and
  Farhadi]{zhu2017target}
Y.~Zhu, R.~Mottaghi, E.~Kolve, J.~J. Lim, A.~Gupta, L.~Fei-Fei, and A.~Farhadi.
\newblock Target-driven visual navigation in indoor scenes using deep
  reinforcement learning.
\newblock In \emph{2017 IEEE international conference on robotics and
  automation (ICRA)}, pages 3357--3364. IEEE, 2017.

\bibitem[Mishkin et~al.(2019)Mishkin, Dosovitskiy, and
  Koltun]{mishkin2019benchmarking}
D.~Mishkin, A.~Dosovitskiy, and V.~Koltun.
\newblock Benchmarking classic and learned navigation in complex 3d
  environments.
\newblock \emph{arXiv preprint arXiv:1901.10915}, 2019.

\bibitem[Anderson et~al.(2018)Anderson, Chang, Chaplot, Dosovitskiy, Gupta,
  Koltun, Kosecka, Malik, Mottaghi, Savva, et~al.]{anderson2018evaluation}
P.~Anderson, A.~Chang, D.~S. Chaplot, A.~Dosovitskiy, S.~Gupta, V.~Koltun,
  J.~Kosecka, J.~Malik, R.~Mottaghi, M.~Savva, et~al.
\newblock On evaluation of embodied navigation agents.
\newblock \emph{arXiv preprint arXiv:1807.06757}, 2018.

\bibitem[Sax et~al.(2020)Sax, Zhang, Emi, Zamir, Savarese, Guibas, and
  Malik]{sax2019learning}
A.~Sax, J.~O. Zhang, B.~Emi, A.~Zamir, S.~Savarese, L.~Guibas, and J.~Malik.
\newblock Learning to navigate using mid-level visual priors.
\newblock In \emph{International Conference on Learning Representations}, 2020.

\bibitem[Wijmans et~al.(2019)Wijmans, Kadian, Morcos, Lee, Essa, Parikh, Savva,
  and Batra]{wijmans2019dd}
E.~Wijmans, A.~Kadian, A.~Morcos, S.~Lee, I.~Essa, D.~Parikh, M.~Savva, and
  D.~Batra.
\newblock {DD-PPO}: Learning near-perfect pointgoal navigators from 2.5 billion
  frames.
\newblock \emph{arXiv}, pages arXiv--1911, 2019.

\bibitem[Chaplot et~al.(2020)Chaplot, Gupta, Gandhi, Gupta, and
  Salakhutdinov]{chaplot2020learning}
D.~S. Chaplot, S.~Gupta, D.~Gandhi, A.~Gupta, and R.~Salakhutdinov.
\newblock Learning to explore using active neural mapping.
\newblock In \emph{International Conference on Learning Representations}, 2020.
\newblock URL \url{https://openreview.net/pdf?id=HklXn1BKDH}.

\bibitem[Yang et~al.(2019)Yang, Wang, Farhadi, Gupta, and
  Mottaghi]{yang2018visual}
W.~Yang, X.~Wang, A.~Farhadi, A.~Gupta, and R.~Mottaghi.
\newblock Visual semantic navigation using scene priors, 2019.

\bibitem[Mousavian et~al.(2019)Mousavian, Toshev, Fi{\v{s}}er,
  Ko{\v{s}}eck{\'a}, Wahid, and Davidson]{mousavian2019visual}
A.~Mousavian, A.~Toshev, M.~Fi{\v{s}}er, J.~Ko{\v{s}}eck{\'a}, A.~Wahid, and
  J.~Davidson.
\newblock Visual representations for semantic target driven navigation.
\newblock In \emph{2019 International Conference on Robotics and Automation
  (ICRA)}, pages 8846--8852. IEEE, 2019.

\bibitem[Zhang et~al.(2017)Zhang, Tai, Boedecker, Burgard, and
  Liu]{zhang2017neural}
J.~Zhang, L.~Tai, J.~Boedecker, W.~Burgard, and M.~Liu.
\newblock Neural slam: Learning to explore with external memory.
\newblock \emph{arXiv preprint arXiv:1706.09520}, 2017.

\bibitem[Chen et~al.(2019)Chen, Gupta, and Gupta]{chen2018learning}
T.~Chen, S.~Gupta, and A.~Gupta.
\newblock Learning exploration policies for navigation.
\newblock In \emph{International Conference on Learning Representations}, 2019.

\bibitem[Parisotto and Salakhutdinov(2017)]{parisotto2017neural}
E.~Parisotto and R.~Salakhutdinov.
\newblock Neural map: Structured memory for deep reinforcement learning.
\newblock \emph{arXiv preprint arXiv:1702.08360}, 2017.

\bibitem[Savinov et~al.(2018)Savinov, Dosovitskiy, and Koltun]{savinov2018semi}
N.~Savinov, A.~Dosovitskiy, and V.~Koltun.
\newblock Semi-parametric topological memory for navigation.
\newblock \emph{arXiv preprint arXiv:1803.00653}, 2018.

\bibitem[Fang et~al.(2019)Fang, Toshev, Fei-Fei, and Savarese]{fang2019scene}
K.~Fang, A.~Toshev, L.~Fei-Fei, and S.~Savarese.
\newblock Scene memory transformer for embodied agents in long-horizon tasks.
\newblock In \emph{Proceedings of the IEEE Conference on Computer Vision and
  Pattern Recognition}, pages 538--547, 2019.

\bibitem[Schulman et~al.(2017)Schulman, Wolski, Dhariwal, Radford, and
  Klimov]{schulman2017proximal}
J.~Schulman, F.~Wolski, P.~Dhariwal, A.~Radford, and O.~Klimov.
\newblock Proximal policy optimization algorithms.
\newblock \emph{arXiv preprint arXiv:1707.06347}, 2017.

\bibitem[Haarnoja et~al.(2018)Haarnoja, Zhou, Abbeel, and
  Levine]{haarnoja2018soft}
T.~Haarnoja, A.~Zhou, P.~Abbeel, and S.~Levine.
\newblock Soft actor-critic: Off-policy maximum entropy deep reinforcement
  learning with a stochastic actor.
\newblock \emph{arXiv preprint arXiv:1801.01290}, 2018.

\bibitem[Zamir et~al.(2018)Zamir, Xia, He, and Sax]{Gibson_CVPR18}
A.~Zamir, F.~Xia, Z.-Y. He, and S.~Sax.
\newblock Gibson environment for embodied real-world active perception.
\newblock \emph{CVPR (to appear)}, 2018.

\bibitem[Savva et~al.(2019)Savva, Kadian, Maksymets, Zhao, Wijmans, Jain,
  Straub, Liu, Koltun, Malik, et~al.]{savva2019habitat}
M.~Savva, A.~Kadian, O.~Maksymets, Y.~Zhao, E.~Wijmans, B.~Jain, J.~Straub,
  J.~Liu, V.~Koltun, J.~Malik, et~al.
\newblock Habitat: A platform for embodied ai research.
\newblock In \emph{Proceedings of the IEEE International Conference on Computer
  Vision}, pages 9339--9347, 2019.

\bibitem[Khatib(1986)]{khatib1986real}
O.~Khatib.
\newblock Real-time obstacle avoidance for manipulators and mobile robots.
\newblock In \emph{Autonomous robot vehicles}, pages 396--404. Springer, 1986.

\bibitem[Fox et~al.(1997)Fox, Burgard, and Thrun]{fox1997dynamic}
D.~Fox, W.~Burgard, and S.~Thrun.
\newblock The dynamic window approach to collision avoidance.
\newblock \emph{IEEE Robotics \& Automation Magazine}, 4\penalty0 (1):\penalty0
  23--33, 1997.

\bibitem[Chen et~al.(2015)Chen, Seff, Kornhauser, and
  Xiao]{chen2015deepdriving}
C.~Chen, A.~Seff, A.~Kornhauser, and J.~Xiao.
\newblock Deepdriving: Learning affordance for direct perception in autonomous
  driving.
\newblock In \emph{Proceedings of the IEEE International Conference on Computer
  Vision}, pages 2722--2730, 2015.

\bibitem[Faust et~al.(2018)Faust, Oslund, Ramirez, Francis, Tapia, Fiser, and
  Davidson]{faust2018prm}
A.~Faust, K.~Oslund, O.~Ramirez, A.~Francis, L.~Tapia, M.~Fiser, and
  J.~Davidson.
\newblock {PRM-RL:} long-range robotic navigation tasks by combining
  reinforcement learning and sampling-based planning.
\newblock In \emph{2018 IEEE International Conference on Robotics and
  Automation (ICRA)}, pages 5113--5120. IEEE, 2018.

\bibitem[Yang et~al.(2019)Yang, Ren, Xu, Chen, Crandall, Parikh, and
  Batra]{yang2019embodied}
J.~Yang, Z.~Ren, M.~Xu, X.~Chen, D.~Crandall, D.~Parikh, and D.~Batra.
\newblock Embodied visual recognition.
\newblock \emph{arXiv preprint arXiv:1904.04404}, 2019.

\bibitem[W{\"o}gerer et~al.(2012)W{\"o}gerer, Bauer, Rooker, Ebenhofer,
  Rovetta, Robertson, and Pichler]{wogerer2012locobot}
C.~W{\"o}gerer, H.~Bauer, M.~Rooker, G.~Ebenhofer, A.~Rovetta, N.~Robertson,
  and A.~Pichler.
\newblock {LOCOBOT}-low cost toolkit for building robot co-workers in assembly
  lines.
\newblock In \emph{International conference on intelligent robotics and
  applications}, pages 449--459. Springer, 2012.

\bibitem[Abadi et~al.(2016)Abadi, Barham, Chen, Chen, Davis, Dean, Devin,
  Ghemawat, Irving, Isard, et~al.]{abadi2016tensorflow}
M.~Abadi, P.~Barham, J.~Chen, Z.~Chen, A.~Davis, J.~Dean, M.~Devin,
  S.~Ghemawat, G.~Irving, M.~Isard, et~al.
\newblock Tensorflow: A system for large-scale machine learning.
\newblock In \emph{12th $\{$USENIX$\}$ Symposium on Operating Systems Design
  and Implementation ($\{$OSDI$\}$ 16)}, pages 265--283, 2016.

\bibitem[Chen et~al.(2016)Chen, Pan, Monga, Bengio, and
  Jozefowicz]{chen2016revisiting}
J.~Chen, X.~Pan, R.~Monga, S.~Bengio, and R.~Jozefowicz.
\newblock Revisiting distributed synchronous sgd.
\newblock In \emph{International Conference on Learning Representations
  Workshop Track}, 2016.

\bibitem[Armeni et~al.(2019)Armeni, He, Gwak, Zamir, Fischer, Malik, and
  Savarese]{armeni20193d}
I.~Armeni, Z.-Y. He, J.~Gwak, A.~R. Zamir, M.~Fischer, J.~Malik, and
  S.~Savarese.
\newblock 3{D} scene graph: A structure for unified semantics, 3d space, and
  camera.
\newblock In \emph{Proceedings of the IEEE International Conference on Computer
  Vision}, pages 5664--5673, 2019.

\bibitem[Kuipers and Levitt(1988)]{kuipers1988navigation}
B.~J. Kuipers and T.~S. Levitt.
\newblock Navigation and mapping in large scale space.
\newblock \emph{AI magazine}, 9\penalty0 (2):\penalty0 25--25, 1988.

\bibitem[Haarnoja et~al.(2018)Haarnoja, Zhou, Hartikainen, Tucker, Ha, Tan,
  Kumar, Zhu, Gupta, Abbeel, et~al.]{haarnoja2018app}
T.~Haarnoja, A.~Zhou, K.~Hartikainen, G.~Tucker, S.~Ha, J.~Tan, V.~Kumar,
  H.~Zhu, A.~Gupta, P.~Abbeel, et~al.
\newblock Soft actor-critic algorithms and applications.
\newblock \emph{arXiv preprint arXiv:1812.05905}, 2018.

\bibitem[He et~al.(2016)He, Zhang, Ren, and Sun]{he2016deep}
K.~He, X.~Zhang, S.~Ren, and J.~Sun.
\newblock Deep residual learning for image recognition.
\newblock In \emph{CVPR}, pages 770--778, 2016.

\end{thebibliography}

\clearpage
\section{Appendix}
\subsection{Soft Actor-Critic}

As commonly done, we formulate the problem as a Partially Observable Markov Decision Process (POMDP) $(\mathcal{O}, \mathcal{S}, \mathcal{A}, P, R)$. The observation set $\mathcal{O}$ consists of sensory measurements such RGB images, depth images, and LiDAR readings (see Sec.~\ref{Sec:observations}). The state space $\mathcal{S}$ captures the current and past poses of the robot and environment. $\mathcal{A}$ is the action space, which in our case consists of twist commands (linear and angular velocities, see Sec.~\ref{Sec:observations}). $P$ represents the state transition probability of the next state $s_{t+1} \in \mathcal{S}$ given the current state $s_t \in \mathcal{S}$ and action $a_t \in \mathcal{A}$. Lastly, $R$ is the reward, which is discussed in detail in Section \ref{sec:reward}.

We learn a navigation policy $\pi(a_t \mid o_t)$ using Soft Actor-Critic (SAC)~\cite{haarnoja2018soft,haarnoja2018app}, an off-policy form of maximum entropy reinforcement learning. The idea behind maximum entropy reinforcement learning is to learn a policy that maximally accomplishes the task while simultaneously being maximally random, with the idea that this randomness aids both exploration and robustness. To accomplish this, SAC learns a policy that maximizes the expected sum of rewards along with the $\alpha$-temperature weighted entropy $\mathcal{H}$ over the policy ~\cite{haarnoja2018soft},
\[
\pi^* = \argmax_{\pi} \mathbb{E}_{\tau \sim \pi}\sum_{t=1}^T R(s_t, a_t) + \alpha \mathcal{H}(\pi(\cdot \mid o_t))
\]
with $\tau=(s_0, o_0, a_1, s_1, o_1,\cdots, a_{T}, s_{T}, o_{T})$ sampled from $\pi$. The above policy is learned in an Actor-Critic setting, where the policy is optimized to be consistent with a critic, a state-action function. The latter is learned via a `soft' Bellman residual optimization. %For further details consult Haarnoja et al.~\cite{haarnoja2018app}.

The optimization iterates between two steps. In the \textit{experience collect} step, sequences of state transitions and associated actions are sampled from the environment using a recent policy. These transitions are used to perform the aforementioned optimization in the \textit{training} step.

\subsection{Implementation Details}\label{sec:impl_details}
\textbf{NN Architecture}
We use the same NN architecture for all function approximators used in the SAC implementation, the policy, state action function, and value function. This architecture consists of observation embedders whose outputs are concatenated and fed into a single layer LSTM of dimension $512$. 

As an observation embedder we use a ResNet50~\cite{he2016deep}, whose channel number is scaled down by a factor of $4$. All image observations, (\texttt{RGB}, \texttt{DEPTH}, \texttt{DET}), are concatenated along their channel dimension and fed into the network. The last observation, \texttt{LiDAR}, is 1-dimensional and is embedded using a 3 layer ConvNet, with a final fully connected layer. All resulting observation embeddings are of dimension $128$ and are added together as a final observation embedding. 

The auxiliary observations, denoting the previous action and its success, are independently embedded using two 2-layer MLP, both layers of dimension $128$.

In addition to the above observation, the policy is conditioned by the label of the target object. This label is represented by a 1-hot vector, which is embedded by a similar 2-layer MLP, both layers of dimension $128$.

\textbf{Training Details} We use the Adam optimizer with a learning rate of 0.000316. The LSTM unroll length during training is $20$, which is substantially smaller than the maximum step length of $500$ during evaluation and of $100$ during collection. Finally, SAC uses lagging weights for the target state action function in its Bellman error loss to stabilize training. In our implementation we gradually update these weights every step with a Polyak update of weight $0.005$ every 1 step. We use a gamma discount factor of $0.99$. We performed a grid search over only the learning rate, in the range $[\textrm{1e-4}, \textrm{1e-3}]$.

\subsection{Evaluation Details}\label{sec:eval_details}

\textbf{Metric Definitions} (SPL)~\cite{anderson2018evaluation}, defined over $N$ navigation episodes, is the average of the success indicators $S_i$  scaled by navigation efficiency, expressed in terms of the optimal path length $o_i$ to goal and the length $l_i$ of the actual path taken: $\textrm{SPL}=\frac{1}{N} \sum_{i=1}^N S_i \frac{o_i}{\max\{l_i, o_i\}}$. We note that a perfect SPL of 1.0 is not realistic to expect because navigation decisions in unexplored environments can be ambiguous.

\textbf{Human Raters}
We provide a virtual setup where a rater can perform `forward', `turn left', `turn right' discrete actions using the keys `W', `A', and `D'. The rater has not seen the environments before, and is allowed a single navigation episode per home. We use the same criteria for starting point and episode length as for the policy evaluation.

% \subsection{Baselines}
% \begin{description}
%     \item[Roomba:] the policy moves straight until collision. When it collides, it picks a random rotation direction, performs rotations until collision free, and repeats.
%     \item[Topo Graph Traversal (TGT):] the policy follows a topological graph of the environment, defined as the skeleton of the floorplan~\cite{kuipers1988navigation}. We perform a depth-first traversal of this graph. This guarantees an efficient coverage of the space, however the space is explored without taking into account the goal label. This baseline is by definition collision free.
% \end{description}

\end{document}